# Closed Curves and Elementary Visual Object Identification

Manfred Harringer [*]

## Abstract

For two closed curves on a plane (discrete version) and local criteria for similarity of points on the curves one gets a potential, which describes the similarity between curve points. This is the base for a global similarity measure of closed curves (Fréchet distance). I use borderlines of handwritten digits to demonstrate an area of application.
I imagine, measuring the similarity of closed curves is an essential and elementary task performed by a visual system. This approach to similarity measures may be used by visual systems.

## Contents

* via researchgate.net

# Introduction

I consider closed curves on a discrete plane. The plane is a lattice with points N x N, N = {0,1, … ,n-1}. The curve is a sequence of neighboured points. I want to present a method to compare such curves by measuring a Fréchet distance [1]. My method contains a specialisation and simplification of the general task. I describe a curve by local features, localized by single points. From there I get a potential, which assigns to each pair of points of two curves the sum of the distances of the local features. I deal with borderlines of digits, especially of handwritten digits available in the MNIST database [3]. In [2] there is a similar task, where Fréchet distances are used.

I imagine, measuring the similarity of closed curves is an essential and elementary task performed by a visual system. I imagine, the visual system always records such curves. Then it is possible to compare recorded curves, which are neighboured in time and mean localization. If high similarity is expected, my method of measuring similarity will work exceedingly well. An elementary object in the environment corresponds to a sequence in time of closed curves in the visual system, where each curve is similar to its successor.

# Similarity of Closed Curves

### *Closed Curves*

I consider closed curves occurring as borderlines of an exemplar of a digit. I get digit exemplars from the MNIST database of handwritten digits [3] available with gray scales from 0 to 255, and 10 hand made digit exemplars '0' to '9', see Appendix. In MNIST, I use the gray scale limit 80 to decide, whether a point belongs to the digit. I search at the border of the digit, i.e. at points belonging to the digit with at least one neighbour outside. I use a simple method to find borderlines:

<u>*start*</u>:
For every border point P there is a suitable successor Q to start with (P,Q). I skip details.
<u>*next*</u>:
If I have two neighboured points P and Q at the border, Q successor of P, I search in the neighbourhood of Q for additional border points by a loop rotating right oriented through all neighbours of Q. Starting at P, the first point <> P, which is a border point, is the successor of Q. At *start* and *next* it is important to fix one orientation: right oriented.
<u>*end condition*</u>:
There occurs a pair (P,Q), which already exists on the curve. In my special case it is sufficient to test, whether there occurs a repetition of the first pair of the curve.

The closed curve is a cyclic (periodic) sequence of points on the plane. The closed curve is like a polygon, therefore I can calculate its orientation. Theses curves are borderlines on a plane. Therefore it is sufficient to distinguish right curves and left curves. May be, there are several borderlines for one digit. I assume, that the digit is connected, and I restrict myself to the borderline with right orientation. This is the borderline, which runs around the digit.

### *Local Features*

I describe each curve by local features, which I derive from the coordinate values width w, height h of a single point, and angles α. Angles are measured as slope of the line from the predecessor to the successor of a point (remark: a triple (P,Q,P) has an angle of 180° in Q). Reliable feature values should be continuous along a curve. Therefore I derive three curves from the original borderline replacing width and height by mean width and mean height of 3, 7 and 13 neighbour points around a center point on the curve. For each point there are five base coordinates (w,h,w+h, w-h, $\alpha$). The features are the (absolute) differences of base coordinates of order 0, 1, 2. Differences at a point are calculated between successor and predecessor. On this way there are $3*5*3 = 45$ features available. There are many methods to select and to modify features. The effect will be low. But some variety on features is important.

### *Normalization*

The MNIST database digits [3] are available on a 28 x 28 plane.
I make some normalizations, before I build the feature values, to simulate the effect of a higher resolution.

*normalizing the distribution of points on the curve*:
The borderline is like a closed polygon. I divide this polygon in 60 parts of the same length. This leads to a curve of 60 points.

*normalizing width and height*:
I assume, that all digits should have the same height. Starting with original coordinates $w(X_i)$ (width) and $h(X_i)$ (height) of a point $X_i$ of the curve X, I derive normalized coordinates $W(X_i)$ and $H(X_i)$.

| | | | |
|---|---|---|---|
| d | := | $\max\{h(X_i)\} - \min\{h(X_i)\}$ | |
| $H(X_j)$ | := | $(h(X_j) - \min\{h(X_i)\}) / d$ | (from 0 ≈ min to 1 ≈ max) |
| $W(X_j)$ | := | $(w(X_j) - \text{mean}\{w(X_i)\}) / d$ | (around the mean value of w, scaled by height) |

### *Comparison of Closed Curves*

I build a free-space-diagram [1]: Two such closed curves build a (discrete) torus, consisting of all index pairs (i,j), where i runs through all indices of points $X_i$ of X, and j runs through all indices of points $Y_j$ of Y. Because of 60 points per curve, the indices run mod 60. The potential V(i,j) at a point (i,j) of the torus is defined by the sum of the absolute differences of the feature values of $X_i,Y_j$ (remark: There are only few features derived from angles. Therefore angle features are weighted by 3, others are weighted by 1.). I call it “potential” because it bears analogy to a physical potential.

The similarity of two curves is measured by Fréchet distances [1]. This is my special approach, based upon the potential:
Low potential values V(i,j) at index pairs (i,j) represent high similarity of the curves X and Y locally in $X_i$ and $Y_j$. Therefore these points $X_i$ and $Y_j$ may be suitable corresponding points of the curves on measuring Fréchet distances. I have to cover all indices i of points $X_i$ on X and all indices j of points $Y_j$ on Y by pairs (i,j), where low potential values V(i,j) are desired. The course of the potential contains valleys of low potential values. If curves X, Y are similar, I expect a deep valley in their potential landscape. Therefore I search for a path through a bottom of a valley. The most obvious procedure is:

<u>*start*</u>:
Find (i,j), where the potential V(i,j) is minimal.
<u>*next*</u>:
The successor candidates of (i,j) are (i,j+1), (i+1,j+1), (i+1,j), where indices run mod 60 because of the normalization. The successor is the candidate, where the potential V is minimal.
<u>*end condition*</u>:
repetition of a pair (i,j)
The resulting curve runs from the first (i,j) to its repetition.
<u>*minmum rule*</u>:
If the minimum is not unique, I decide randomly.

*next* contains a compromise: I have to go forward, but guided by the potential.
The resulting cyclic path through the torus is named ***canonical path*** of the curves X and Y. The mean potential along the canonical path (with the same weight for each point of the path) is my measure of similarity for the curves X and Y. The result is independent of the choice of index 0. Only the successor relation between points on these curves is relevant.

<u>*alternative I*</u> at *start*:
At *start* I select several points, for instance one row. This may result in several cyclic paths through the torus. The canonical path is the path with minimal mean potential (There are examples, where this path does not run through the point with minimal potential.).

<u>*alternative II*</u> at *start* and *next*:
For each point (i,j) on the torus I build a minimal sum of potential values of all possible continuations of prescribed length n with the continuation candidates (i,j+1), (i+1,j+1), (i+1,j) of (i,j) (indices mod 60), recursively over n in $\mathbb{N}_0$. Then I use these sum values instead of the potential values to select the first point at *start* and the successor in the *next*-step. The original methods *start* and *next* correspond to n = 0 in this *alternative II*.

*alternative II* with large n in $\mathbb{N}_0$ may supersede the explicit construction of the canonical path. But there are some obstacles.

## Examples

I compare digit exemplars of the MNIST database of handwritten digits [3] (of the first 1000 digits, enumerated from 0 to 999) with 10 self made patterns for the ciphers '0' to '9'. These 10 patterns are shown in Appendix. Here I show the borderlines, their potential at index pairs (i,j) (60 x 60, i from left to right, j from top to bottom), their canonical path, and the similarity measure as the mean potential at the canonical path.

Low potential values are dark green, high potential values are gray.
The potential values along the canonical paths are recolored for better recognition of small differences.

## *figure I*

MNIST[568], pattern '6'
This is an example for high similarity and good accordance.

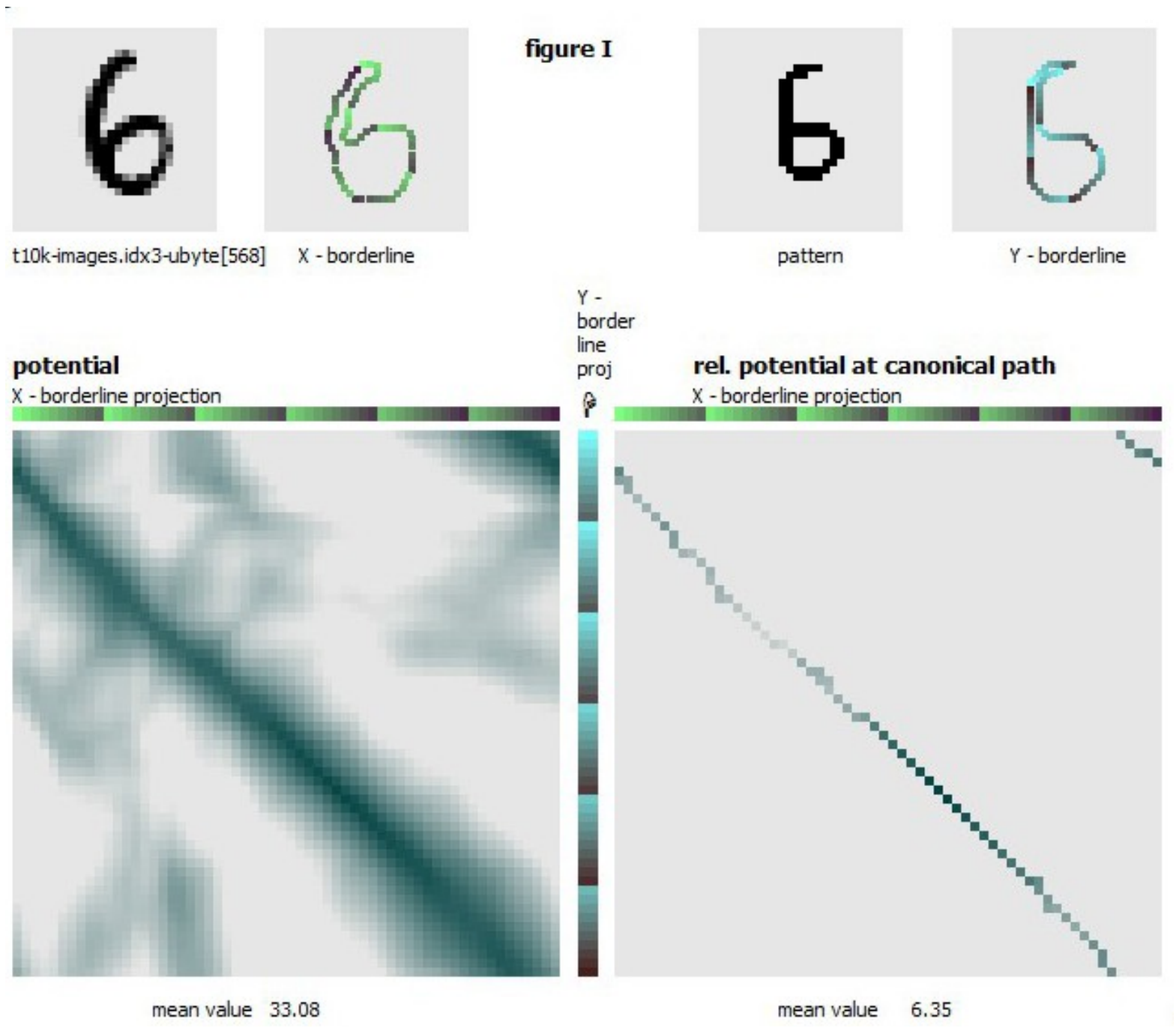

## *figure II*

MNIST[792], pattern '3'
Corresponding kinks are well recognized.

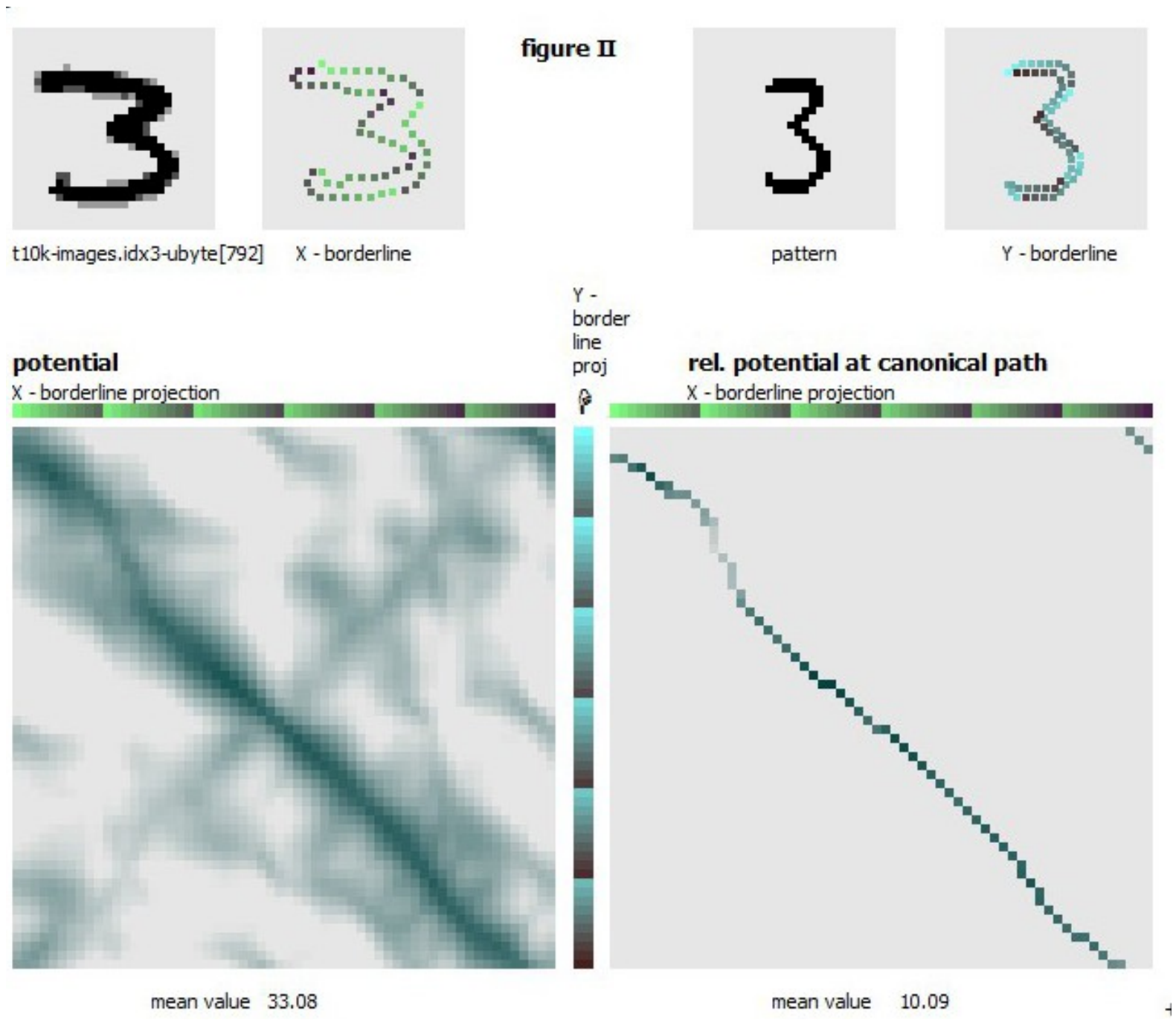

## *figure III and IV*

MNIST[654], pattern '5' and '3'
The best matching patterns for MNIST[654] are '5' and '3'. For pattern '3' a reasonable canonical path is found, with some irritation around the index pair (50,50).

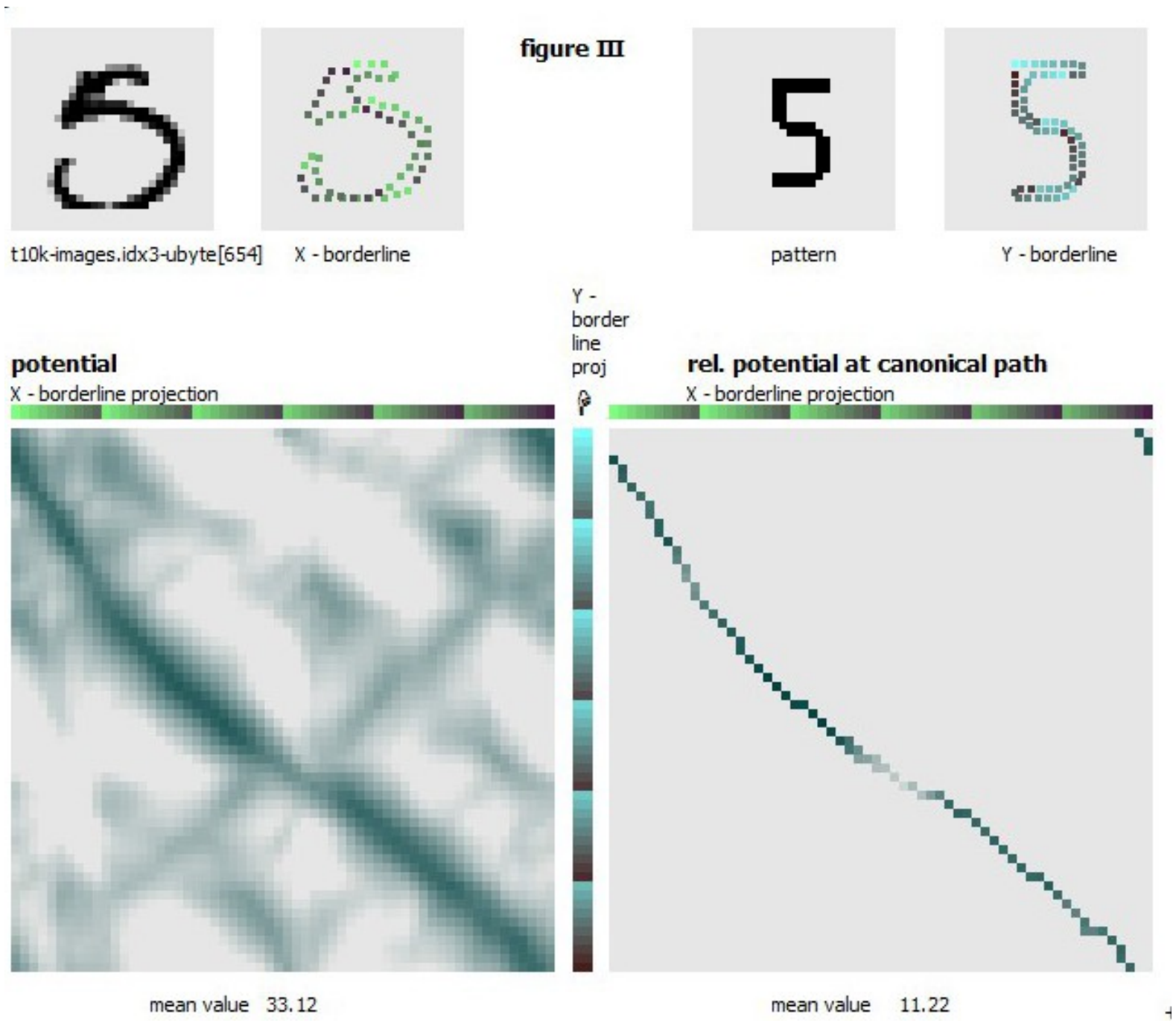

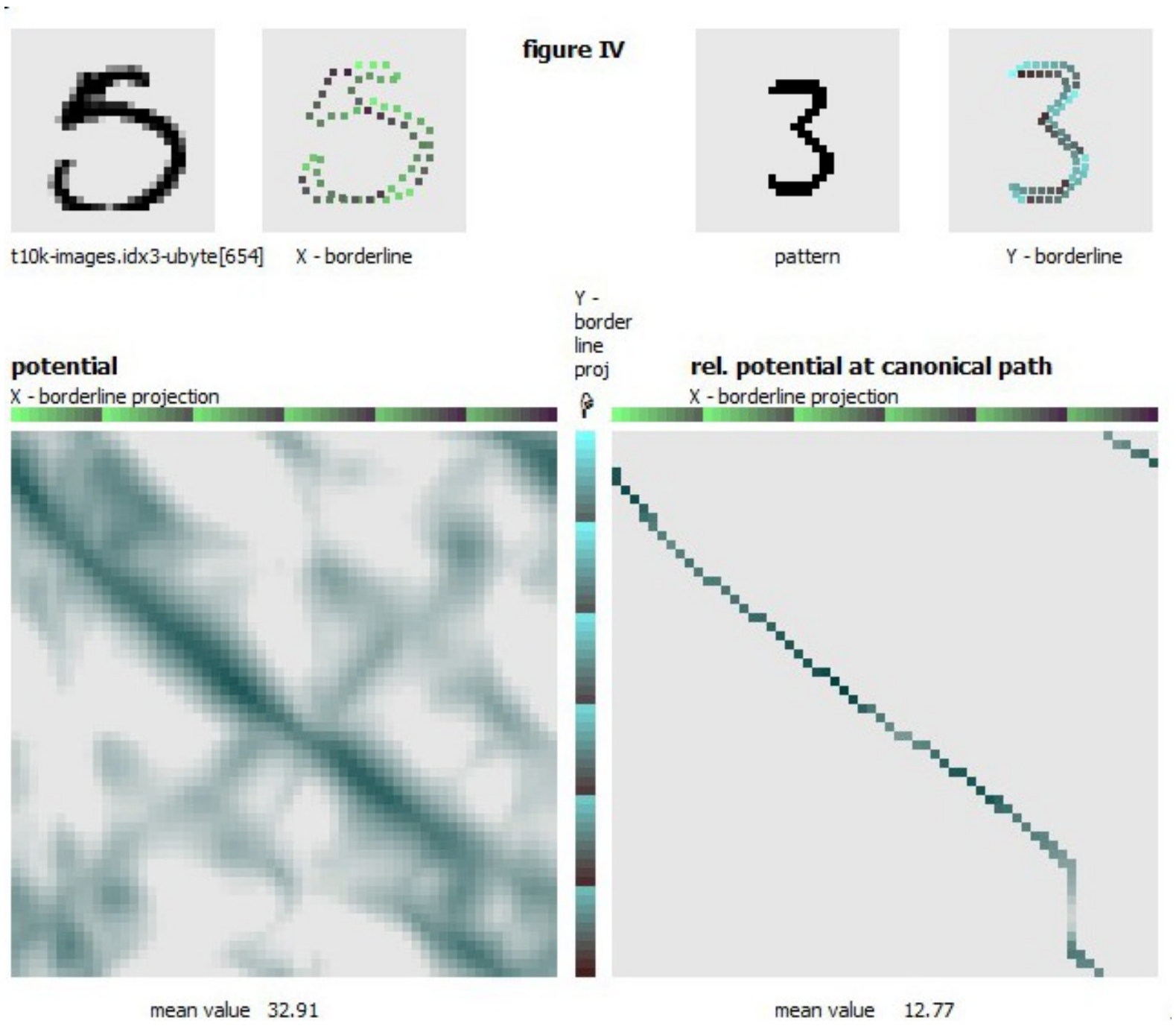

## *figure V*

MNIST[338], pattern '0'
This MINST digit '8' is most similar to the pattern of '0'. This decision is difficult, because I look at one borderline only.

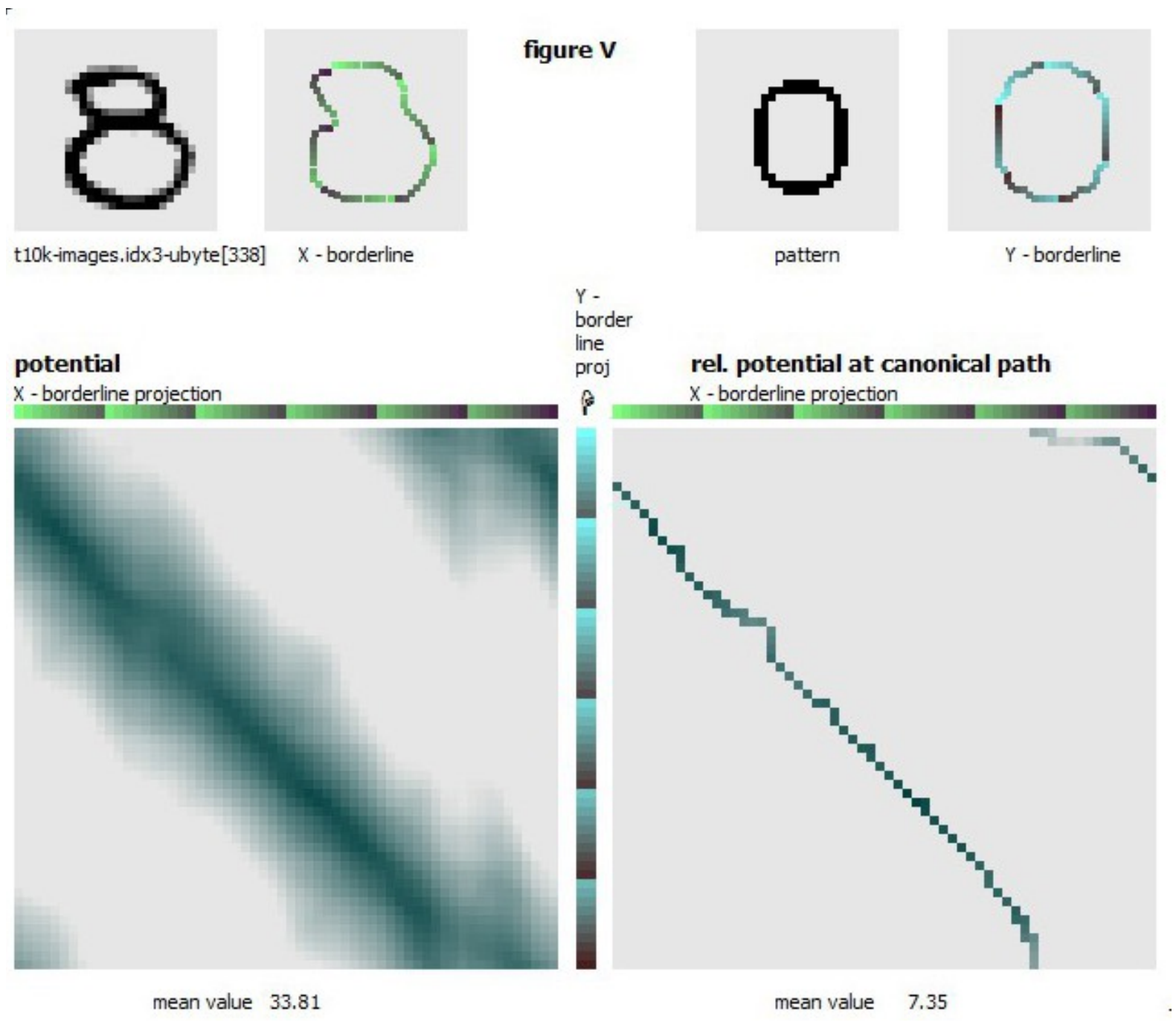

## *figure VI and VII*

MNIST[823], pattern '2', where figure VII uses *alternative II* with n = 20
figure VI shows, that the path is misguided around the indices 12 to 15 by a local valley, which will come to an end and forces a passage through high potential values. *Alternative II* is useful to avoid such misguidances. In this special case a meliorated canonical path is found by propagation in opposite direction, too. But nevertheless the best matching pattern is '7'. The mean potential at the canonical path is always > 13, therefore no pattern is really matching.

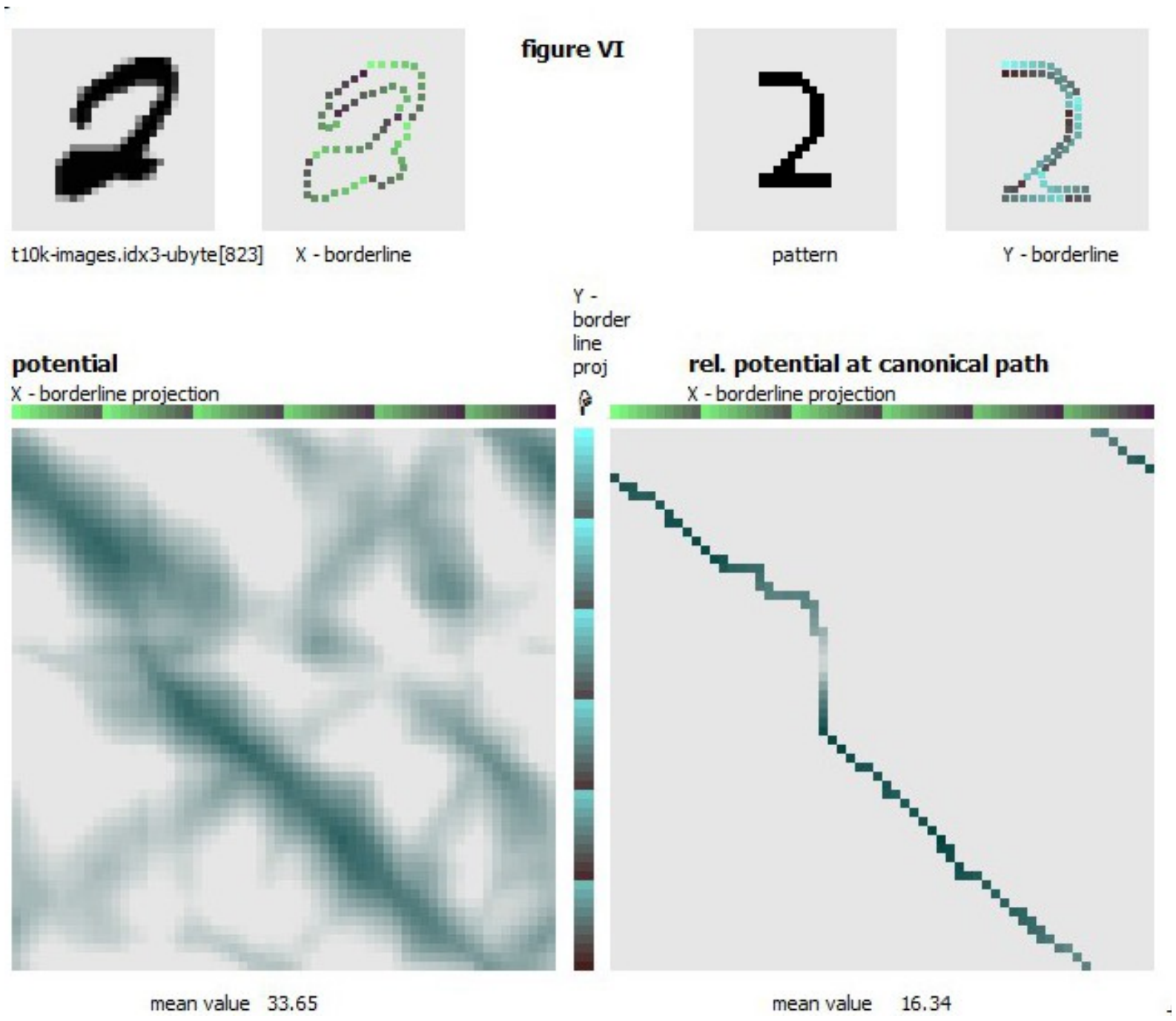

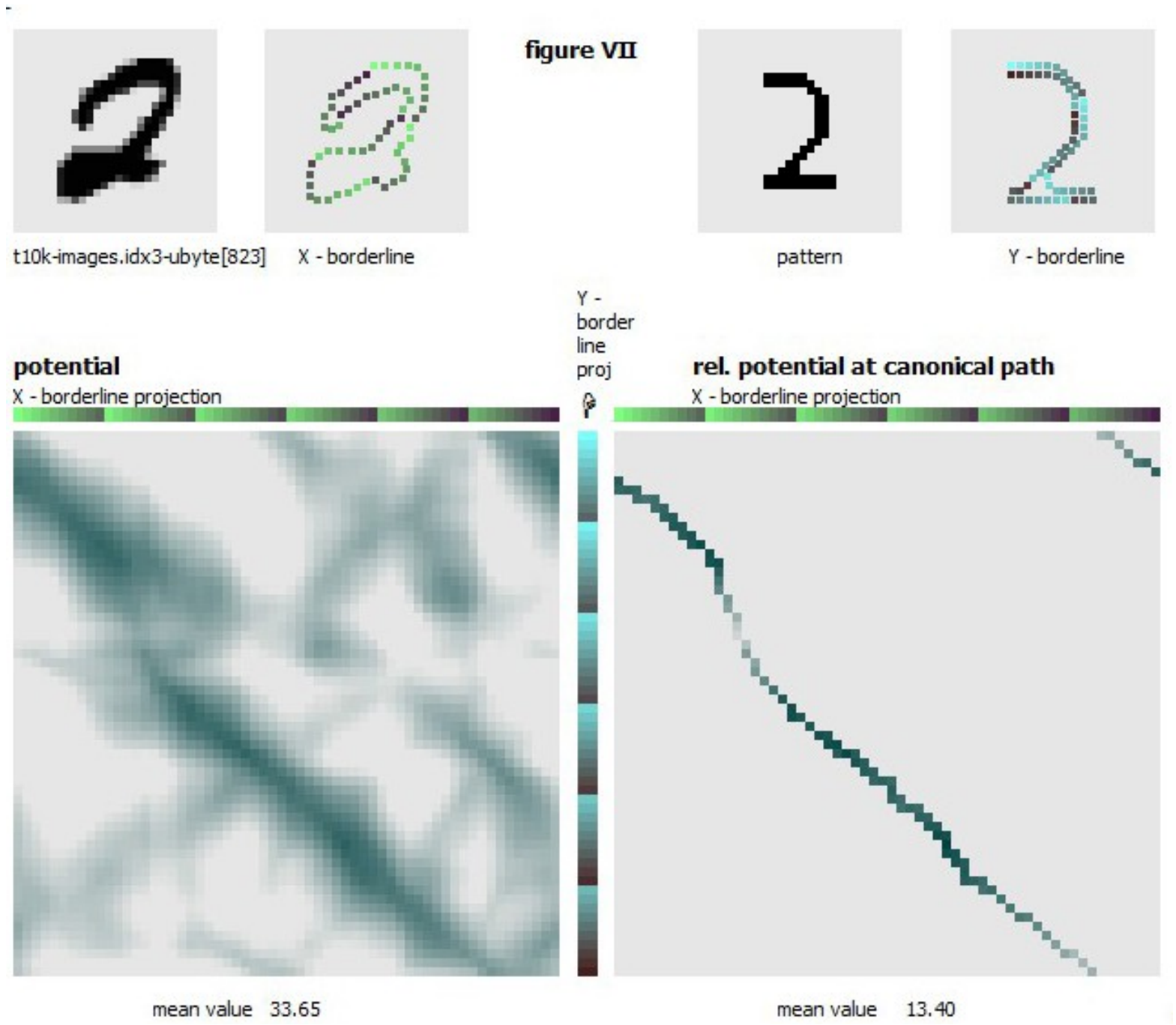

## *figure VIII*

MNIST[671], pattern '9'
Because of the gap in the MNIST digit exemplar of '9' the canonical path has two cycles of the MNIST digit per 1 cycle of the pattern. In that case I expect a high mean potential at the canonical path.

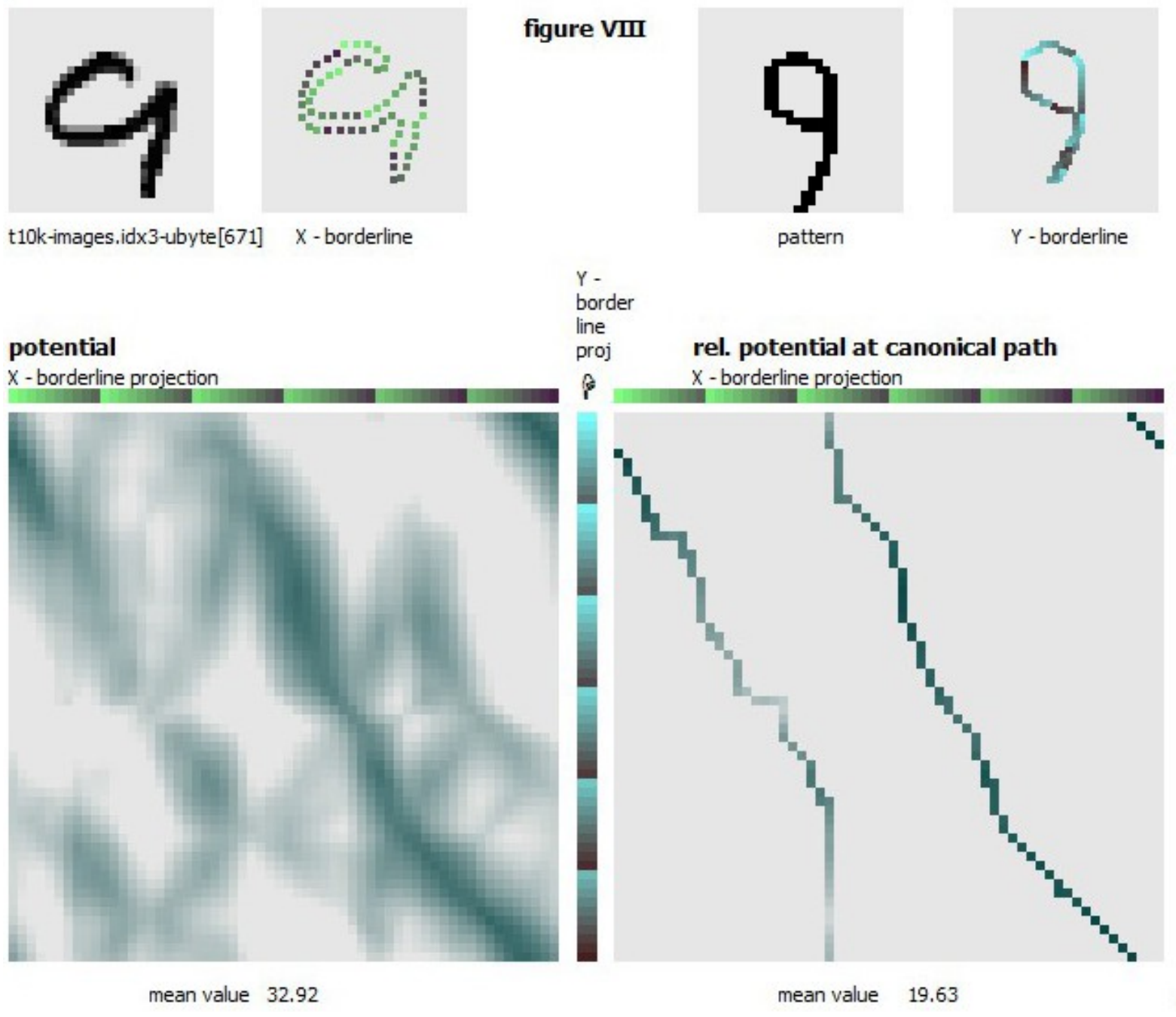

## *figure IX and X*

MNIST[247], *alternative II* with n = 20, pattern '4' and pattern '6'
Pattern '4' is not recognized, pattern '6' is optimal. But the high mean value 13.74 with pattern '6' indicates, that there is no matching pattern.

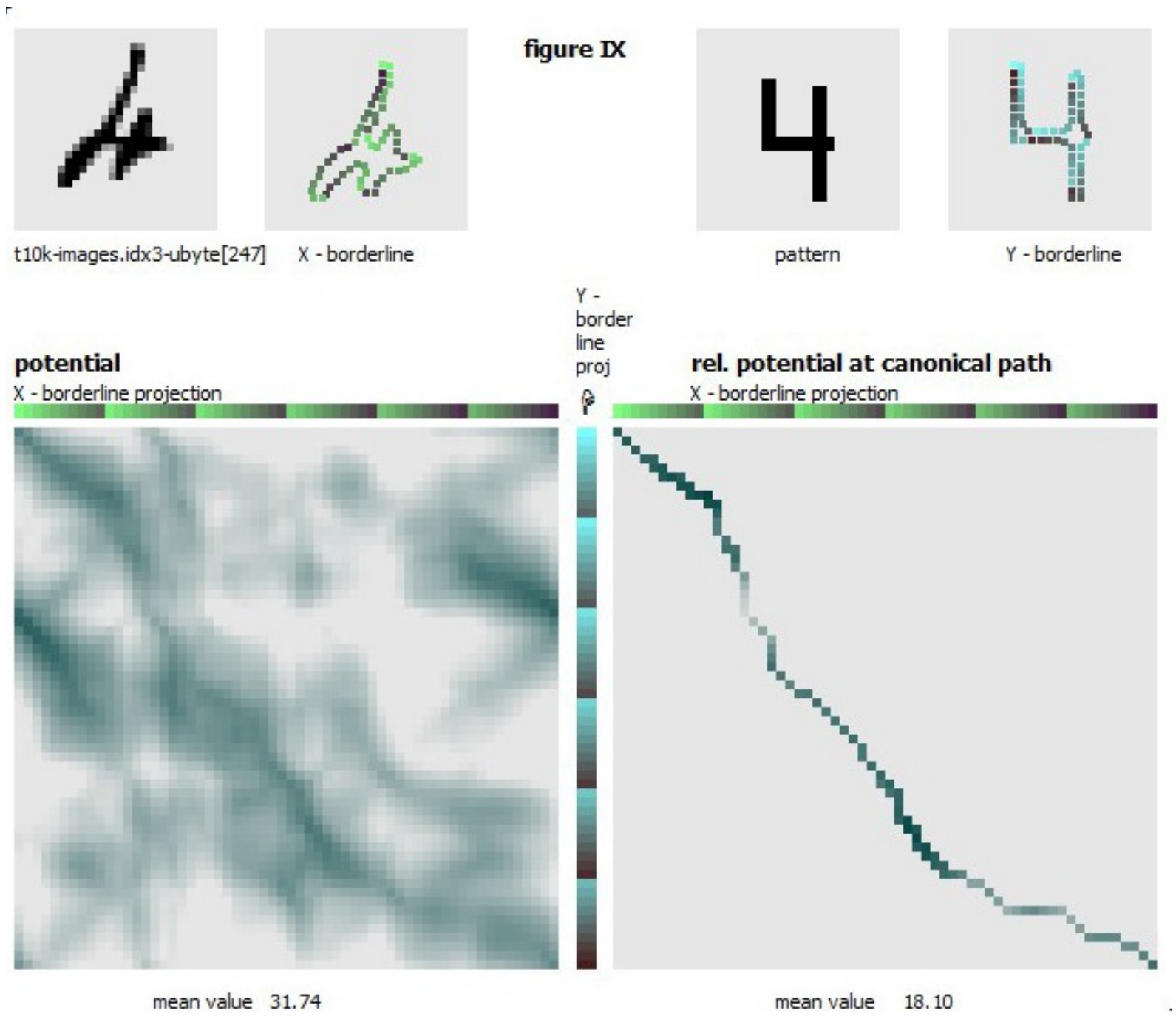

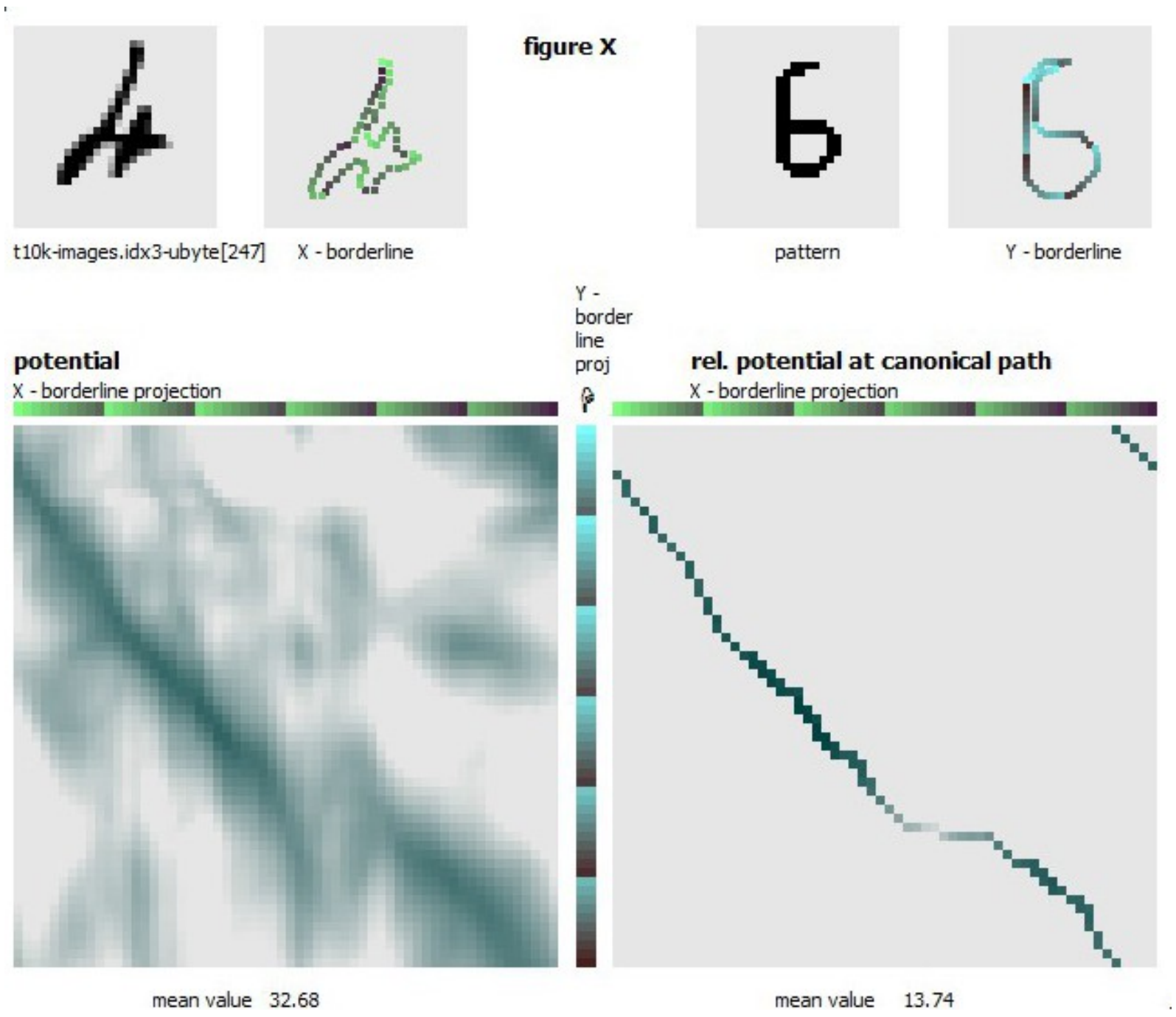

## Variations of the Similarity Measure

### *The Test Series*

I compare the digit exemplars and their meaning of the MNIST database of handwritten digits [3] from number 0 to 999 with 10 self made patterns for the ciphers '0' to '9'. These 10 patterns are shown in Appendix. I calculate the similarity measure as the mean potential at the canonical path. I expect, that the similarity measure of an MNIST digit is minimal for the corresponding pattern in all 10 patterns. In that case the similarity measure makes a true decision, otherwise a false decision. Here I skip a closer look to these decision results (small differences, high values of the similarity measure).

I start with the similarity measure described above. Then the result of the test series is:

1000 MNIST-digits, thereof
- 872 true decisions, therefore
- 128 false decisions, thereof on the following reasons:
  - 45 unexpected gap (example in figure VIII)
  - 11 no gap, where a gap is expected
  - 3 not connected
  - 19 the '2' of an alternate type (example in figures VI, VII)
  - 14 '8' is most similar to '0' (example in figure V)
  - 36 other reasons (example in figures IX, X)

### *Variations*

I present examples with selected features. One kind of features is based upon height, width, (height+width) and (height-width), i.e. “height and width”. The second kind of features is based upon angles.

| selected features | alternative II | true (of 1000) |
|---|---|---|
| all | no | 872 |
| height and width, no differences | no | 610 |
| height and width differences of order 1 | no | 614 |
| height and width differences of order 2 | no | 690 |
| height and width and their differences of order 1, 2 | no | 856 |
| angles, no differences | no | 426 |
| angle differences of order 1 | no | 500 |
| angle differences of order 2 | no | 462 |
| angles and their differences of order 1, 2 | no | 834 |
| all | yes, n = 60 | 888 |
| height and width and their differences of order 1, 2 | yes, n = 60 | 879 |
| angles and their differences of order 1, 2 | yes, n = 60 | 873 |

## Interpretation

### *Character Recognition*

My method, without further refinements, is not suitable for recognition of handwritten digits. Digits consist of lines, but not of borderlines. Neural networks are more suitable [4]. But there is one advantage of my method: I use few patterns for learning, whereas neural networks need many examples. I expect, that my method is suitable for recognition of printed characters.

### *Features*

In the layer visual cortex V1 there are many 2-dimensional charts for orientation and ocular dominance [5]. Ocular dominance is useful in a 3-dimensional context. But my digits are 2-dimensional only. In the visual cortex there is no obvious source for the features, which I use.
remark: When I compare curves neighboured in time and space, the features used there don't need invariances.

### *Closed Curves*

It is well known, that our visual system detects borderlines [6]. Closed curves are an important special case [7]. I use an ad hoc method to find closed curves, probably without biological background.

### *2-Dimensional Charts*

There are functional areas in the visual cortex V1, organized like 2-dimensional charts, used for orientation and ocular dominance [5]. But there are additional 2-dimensional charts, which perform unknown tasks [8]. I pointed out, that this structure occurs in the task of measuring similarity of closed curves by a canonical path through a potential. There are many similar tasks. The curves may be open. More general, if there are sequences with local features, then I can compare two sequences by a 2-dimensional potential, too.

### *Objects*

A visual system needs selection rules to select relevant features for any object identification. For instance the detection of individual human faces starts with the insight, that there is a face [9]. I imagine, the first step is the detection of its borderline. Therefore I claim, that the recognition of closed curves is elementary.

### *Object Tracking*

Methods of object tracking assume, that there is already an object identified and separated and to be traced [10]. I imagine, the visual system detects a closed curve. Then it searches for a chain of similar closed curves. Such a chain, observed for some time, represents an object. In this context the tasks of „identification“ and „tracking“ coincide.

## Appendix

10 self made patterns

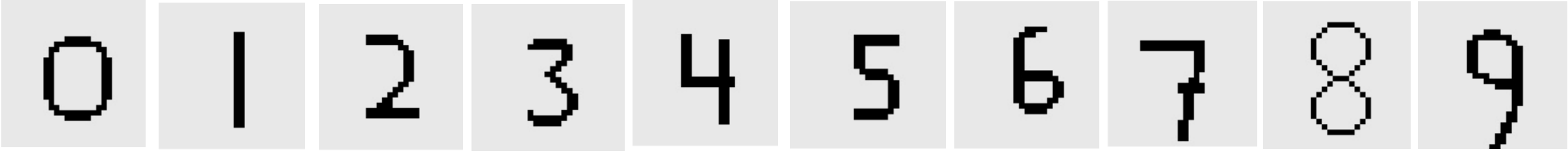